\documentclass[manuscript,screen,9pt,]{acmart}
\usepackage{graphicx} 

\usepackage{comment}
\usepackage{hyperref}
\usepackage{array}
\usepackage{multirow}

\AtBeginDocument{%
  \providecommand\BibTeX{{%
    \normalfont B\kern-0.5em{\scshape i\kern-0.25em b}\kern-0.8em\TeX}}}

\setcopyright{acmlicensed}
\copyrightyear{2024}
\acmYear{2024}
\acmDOI{XXXXXXX.XXXXXXX}

\acmConference[ICISDM2024 '8]{International Conference on Information System and Data Mining}{June 24th-26th, 2024}{Los Angeles, USA}
\begin{document}

\title{A Few-Shot Learning Focused Survey on Recent Named Entity Recognition and Relation Classification Methods}

\author{Sakher Khalil Alqaaidi}
\affiliation{%
  \institution{sakher.a@uga.edu, School of Computing, University of Georgia}
  \city{Athens}
  \state{Georgia}
  \country{USA}
}
\author{Elika Bozorgi}
\affiliation{%
  \institution{elika.bozorgi@uga.edu, School of Computing, University of Georgia}
  \city{Athens}
  \state{Georgia}
  \country{USA}
}
\author{Afsaneh Shams}
\affiliation{%
  \institution{afsaneh.shams@uga.edu, School of Computing, University of Georgia}
  \city{Athens}
  \state{Georgia}
  \country{USA}
}
\author{Krzysztof Kochut}
\affiliation{%
  \institution{kkochut@uga.edu, School of Computing, University of Georgia}
  \city{Athens}
  \state{Georgia}
  \country{USA}
}

\renewcommand{\shortauthors}{}

\begin{abstract}
Named Entity Recognition (NER) and Relation Classification (RC) are important steps in extracting information from unstructured text and formatting it into a machine-readable format. We present a survey of recent deep learning models that address named entity recognition and relation classification, with focus on few-shot learning performance. Our survey is helpful for researchers in knowing the recent techniques in text mining and extracting structured information from raw text.

Our work is a good introduction to any beginner researcher in the two tasks.
\end{abstract}




\begin{teaserfigure}
  \includegraphics[width=\textwidth]{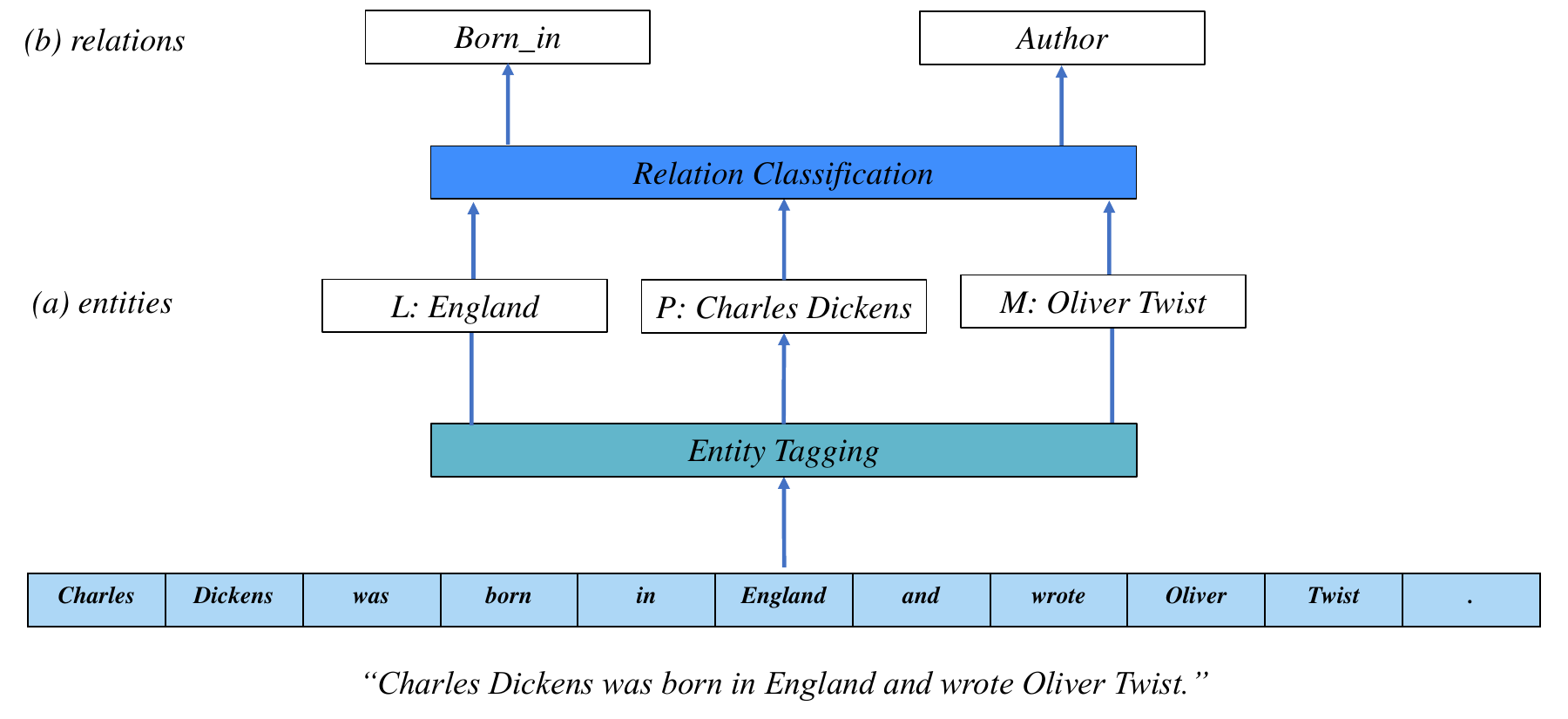}
  \caption{Example of tagging entities and their types then identifying their relations.}
  \Description{Example of tagging entities and their types then identifying their relations.}
  \label{fig:teaser}
\end{teaserfigure}

\received{20 March 2024}

\maketitle
\pagestyle{plain}
\pagenumbering{gobble}

\section{Introduction}
Named Entity Recognition (NER) and Relation Classification (RC) are important steps in extracting information from unstructured text and formatting it into a machine-readable format. Several Natural Language Processing (NLP) applications employ the two steps, either separately or simultaneously, such as information retrieval, knowledge graph construction and completion, question answering and other domain-specific applications, such as biomedical data mining \citep{quirk2016distant}.

The NER task targets labelling subsets of words in text that designate entities. An entity may contain multiple words. Formally, for a sequence of words $W$ of size $n$, $W = \{w_1, w_2 ... w_n \}$, where $w$ is a word in the sequence, entity tagging targets learning the function $f(W) = E$, where $E$ is a set of one or more $e$ entities; $e \subset W$. An entity $e$ may contain multiple $w$ words. It is not necessary that all words within an entity are adjacent, this type of entities is called a discontinuous entity. For example, the term \textit{``The teams of France and Italy"} incurs two entities \textit{``The team of France"} and \textit{``The team of Italy"}. An entity of multiple words may contain instances of sub entities. For example, \textit{``The governor of Bryxton"} is an entity, and \textit{``Bryxton"} is a sub entity; this type of entities is called nested entities. The NER task does not only tag entities in text but also classifies each item into one type of a predefined set of entity types, such as Person, Location, Organization, etc.

The RC task aims to identify if a relation exists between two given entities and classifying the relation into one of predefined semantic relations given in the input. Formally the RC task is defined as:

\begin{equation}
    f(W, E, P) = \left\{ \begin{array}{ll}
         R, & \textrm{one or more valid relations} \\
         \emptyset, & \textrm{otherwise}
    \end{array}
    \right.
    \end{equation}

where $W$ is a sequence of words $\{ w_1,\ w_2\ ...\ w_n \}$, $E$ is a set of one or more entity pairs;  $E = \{ (e_1,e_2)_1 .. (e_1,e_2)_k \} $. Each entity pair consists of a subject entity and an object entity, where an entity $e$ is a sub-sequence of $W$, the entity pair can be defined as a tuple $(e_1, e_2)$. $R$ is a set of one or more relations found for $E$. If the relations are directed, then $e_1$ would be a head or subject entity and $e_2$ would be a tail or object entity. $\emptyset$ indicates that no relation exists in any entities pair. $P$ is a set of predefined semantic relations. Relation Extraction (RE) is known in the literature as the task that incorporates both the NER and RC tasks. Figure \ref{fig:teaser} shows an example of using the two tasks to extract entities and relations from a sentence; At step (a), the entities are tagged along with their types in the first upper case letter. At step (b), the relations are being identified for entities pairs. For instance, ``\textit{Born\_in}'' was found to be a valid relation for the subject entity \textit{``Charles Dickens''} and the object entity \textit{``England''}.

In this work we present a survey on recent approaches in the NER and RC tasks with focus on few-shot learning approaches. Early methods in both tasks used rule-based algorithms, i.e., non-machine learning methods, such as text pattern mining \citep{huffman1995learning}, feature-based methods \citep{kambhatla2004combining} or graphical methods \citep{mcclosky2011event}. Followed that, models that used text representation in neural networks. In this survey, we only consider machine learning-based models for the following reasons: pattern-based or feature-based models have significant lower scores on several benchmarks compared to deep learning models according to several surveys \citep{nasar2021named,yadav2019survey}. Additionally, a few models adopted pattern-based or feature-based methods solely during the last few years, reasoned by the lower scores. Furthermore, several surveys have addressed non-machine learning models sufficiently.

Although supervised learning models have achieved astonishing results in the NER and RC tasks, they suffer from lower accuracy in some practical scenarios. That is when data has no labels or has a few samples labelled. Primitively, this issue was handled by weak or distant supervision models \citep{hoffmann2011knowledge}. However, noisy labels have always put obstacles in reaching good results in the weak or distant supervision models. Thus, there was a need to adopt a new learning approach that can tackle the previously mentioned obstacles. Few-shot learning is a branch of meta-learning, that conducts training on few labelled data and uses a small support set to perform predictions \citep{bragg2021flex}. Few-shot learning has shown remarkable performances in several NLP tasks including NER and RC. Furthermore, Few-shot learning models can easily adapt to various domains with satisfactory results due to its ability to use a few samples and handle labels that were not seen during training. These reasons make this discipline more close to real-world scenarios. Thus, We focus our selection of models in this survey on few-shot learning methods in addition to supervised learning models. We show our methodology of selecting works for this survey in Section \ref{sec:methodology}.


Recent surveys focused on deep learning models and a few considered surveying both NER and RC approaches in a single study. Our work is the first work that considers the two tasks with focus on few-shot learning methods. The surveys in \citep{yadav2019survey,li2020survey} considered the NER task methods only, they showed early approaches and focused on deep learning models. The survey in \citep{nasar2021named} considered works from the NER and RE tasks with focus also on deep learning models. The survey in \citep{han2020more} reviewed the works in the RE task and categorized them based on their approaches, then discussed more paths in the RE task to be explored.

Our survey is divided as the following: Section \ref{sec:datasets} describes the datasets that we found commonly used in both the NER and RC tasks. Section \ref{sec:methodology} explains our methodology in selecting the models for this survey. Section \ref{sec:unified} shows the models that have been found handling both the NER and RC tasks. \ref{sec:NER} shows the models that solely addressed the NER task. Section \ref{sec:relation} shows the models that addressed the relation classification task only. Finally, we conclude our observations in Section \ref{sec:conclusion}.


\section{Benchmarks} \label{sec:datasets}

In this section we report multiple popular benchmarks in the NER and RC tasks. Statistics about the datasets are shown in Table \ref{tab:benchmarks}.

\subsection{Named Entity Recognition Benchmarks}

\begin{itemize}

\item \textbf{CoNLL2003} \citep{sang2003introduction} or Conference on Computational Natural Language Learning corpus is a named entity recognition dataset, the English version was built using Reuters news corpus. The dataset has four entity types: Persons, Locations, Organizations and Miscellaneous.

\item \textbf{OntoNotes5.0} \citep{weischedel2013ontonotes} is an annotated text dataset that has part of speech (POS) and NER tags, built on a corpus of various types of text content, such as news, conversational telephone speech, weblogs, newsgroups, broadcast and talk shows. OntoNotes has different language variants including English.

\item \textbf{FEW-NERD} \citep{ding2021few} is the first released dataset for few-shot NER. Before its release, models that targeted to evaluate their work on the few-shot performance used datasets designed for supervised-learning then customized them for few-shot testing. These customizations led to inconsistent comparison and added difficulties when employing the datasets for few-shot learning due to the variety of entity types and quantities. Thus, FEW-NERD has given a realistic evaluation of models performance on few-shot learning since it is constructed specifically for this task. Particularly, the dataset presented two variants. In FEW-NERD (INTRA), the evaluation entity types are not seen during training, which makes it harder compared to FEW-NERD (INTRA), where splits share the same types. The dataset sentences were retrieved from Wikipedia articles. The dataset has 491.7k annotated entities of 8 coarse-grained types and 66 fine-grained types.

\end{itemize}

\subsection{Relation Classification Benchmarks}


\begin{itemize}
    \item \textbf{TacRed} \citep{zhang2017position} is a dataset that has been used in both the RC and RE tasks. It is derived from news articles and web content. The original release had 41 relation types. The dataset has been designed for supervised learning evaluation. The work in \citep{sabo2021revisiting} showed some drawbacks in popular few-shot learning datasets and proposed an approach to customize the supervised learning ones for few-shot evaluation, such as TacRed. Later on, Re-TacRed was released as an improved version of the original one \citep{stoica2021re}.
    
    \item \textbf{FewRel} \citep{gao2019fewrel} is a Few-Shot relation classification dataset of 100 relations in sentences derived from Wikipedia and labelled by crowdsourcing. The training part has 64 relations, the validation part has 16 relations and the test part has 20 relations. Soon after the release of FewRel, authors presented a new version to examine the models' ability to adapt to new domains. Although FewRel was adopted by many works, the study in \citep{sabo2021revisiting} showed that the dataset is still far from real-world scenarios, thus authors proposed a mechanism to switch supervised datasets, such as TACRED, to be applicable to the few-shot training.

\end{itemize}

\subsection{Relation Extraction Benchmarks}

\begin{itemize}
    \item \textbf{NYT} \cite{riedel2010modeling} is a dataset that was generated from a large New York Times articles corpus, where each input item consisted of a sentence and a set of triples, each triple is composed of subject and object entities, and a relation.
    
    \item \textbf{WebNLG} is a dataset that was originally generated for the Natural Language Generation (NLG) task, CopyRE \cite{zeng2018extracting} customized the dataset for the triples and relations extraction tasks.
\end{itemize}

\begin{table}[t]
    \centering
    \begin{tabular}{lcccc}
    \hline
    Model & Train & Validation & Test & Total\\
    \hline
         CoNLL2003 & 14,041 & 3,250 & 3,453 & 20,744\\
         OntoNotes5.0 & 59,924 & 8,528 & 8,262 & 76,714\\
         FEW-NERD (INTRA) & 99,519 & 19,358 & 44,059 & 162,936\\
         FEW-NERD (INTER) & 130,112 & 18,817 & 14,007 & 162,936\\
         \hline
         TacRed & 68,124 & 22,631 & 15,509 & 106,264\\
         Re-TacRed & 58,465 & 19,584 & 13,418 & 91,467\\
         FewRel & 44,800 & 11,200 & 14,000 & 70,000\\
         \hline
         NYT & 56,196 & 5,000 & 5,000 & 66,196\\
         WebNLG & 5,019 & 500 & 703 & 6,222\\
         \hline
    \end{tabular}
    \caption{Statistics of popular NER, RC, and RE Benchmarks.}
    \label{tab:benchmarks}
\end{table}

 \section{Methodology} \label{sec:methodology}
With hundreds of works in the NER and RC tasks available in the literature and to present a survey that focuses on deep learning-based models for the reasons mentioned in the introduction, we choose the models that were published in 2019 and later; we select this year since it witnessed the beginning of using some revolutionary pre-trained language models (PLMs), such as BERT \citep{devlin2018bert} and GPT \citep{brown2020language}; such PLMs were employed to score new state-of-the-art performances in most of the NLP tasks. With the adoption of English language for many NLP benchmarks and evaluations, we exclude works that pursue other languages solely from our search results. Furthermore, we exclude domain-specific works to survey general-use models that can be adapted for other domains. We searched Google scholar for the terms: \textit{relation extraction}, \textit{named entity recognition} and \textit{relation classification}. We select the papers that has any of the terms in the title or the content that appeared in the first 100 search results, then we give a rank based on the following factors:

\begin{itemize}
    \item Number of citations.
    \item The model presents a few-shot learning results.
    \item The model handles both NER and RC tasks together.
    \item Publication year.
\end{itemize}

The last factor is considered for fairness with papers that were published in the same year of writing this survey and did not receive adequate number of citations



 
\section{Unified NER and RC Models}\label{sec:unified}

In this section we present the models that handled both the NER and RC tasks; the output of these models consisted of either separate entities set and relations set, or joint entities and relations in the form of triples. A triple consists of a subject entity, an object entity and the connecting relation. Some works call the simultaneous NER and RC as the relation extraction (RE) task.  In a sentence, multiple triples may share a single entity in a case named \textit{Single Entity Overlap}, Figure \ref{fig:example} shows an example of the entity \textit{Charles Dickens} that is found in two triples because it is a part of two input items in the RC task. A more complicated scenario when multiple relations connect the same entities, this case is called \textit{Entity Pair Overlap}. For instance, the entities \textit{Bern} and \textit{Switzerland} can have the two relations \textit{capital\_of} and \textit{city\_in} in the sentence \textit{Bern is not only a city in Switzerland but also the capital}.

Early RE models utilized a pipeline approach, where NER or RC is conducted at the beginning then the output is used for running the second task. For instance, entities are extracted first, then used as input in the RC task. However, studies showed that errors from the first stage propagate to the second one and affect the overall performance. Thus, recent models performed a simultaneous validation while training the model.

DeepStruct \citep{wang2022deepstruct} is a supervised learning model with a zero-shot learning variant. Authors showed that language models need structured understanding of text instead of independent aspects like in GPT-3 and BERT. Thus, they proposed to train language models to predict triples as they convey rich information for several NLP tasks, then to utilize multi-task training for downstream tasks including NER, RE and RC. In the zero-shot variant, adequate data was used for the framework training and some datasets were excluded and used for the downstream tasks. LUKE \citep{yamada2020luke} is a BERT-based pretrained language model that utilized entity information in text to achieve better word representation valid for several NLP tasks. Authors followed masking and self-attention approaches different from BERT, which helped recognizing entities. LUKE was tested on different NLP benchmarks including supervised NER performance. The work in \citep{liu2022autoregressive} represented text as actions to build a structure of dependencies between words for supervised learning. The model used T5 language model to encode text. Similar to BERT, T5 is a masked language model. The paper did not mention the approach's ability to handle nested entities. PL-Marker model \citep{ye2021packed} used markers in the text sequence to tag and classify entities and extract entities pair relations. The model considered the neighboring entity spans and subject entities when using the markers. Authors adopted multiple BERT variants for different datasets which weakened the consistency of the evaluation. However, the model supported nested entity tagging. Set Prediction Network (SPN) \citep{sui2023joint} model targeted extracting triples of entities and relations. The model generated a set of triples without going through separating the stages of entity tagging and RC. The model used BERT to encode the text and a novel architecture of a non-autoregressive decorder. Authors proposed a loss function to handle the prediction format of triple sets. The model handled the entities overlapping problems. Authors mentioned the limitation of imbalanced relation distribution in different datasets, which harnessed the model's performance. PURE \citep{zhong2020frustratingly} is a supervised learning model of two components. Initially, the model tagged the entities then used this information for the second stage of relation extraction. Although, the model is simple, errors in the first stage are propagated to the relation extraction level because the first stage output is not validated based on the final output, which is a major defect that was addressed by other models through a joint architecture. Thus, tackling this issue in PURE may boost the performance of the model but will require major changes in the design. The reported results were based on different BERT variants for text encoding.
\renewcommand{\arraystretch}{2}
\begin{table}[t]
    \centering
    \begin{tabular}{p{10em}p{5em}ccccc}
    \hline
         Model & \shortstack{Learning\\Type} & Sent./Doc. &        \shortstack{Nested\\Entities} &      \shortstack{Language\\Model} &    CoNLL03 &         OntoNotes5.0 \\
         \hline
         \citep{yu2020named} &              Supervised &  Sent. &                 Yes &                    BERT &                       93.5 &                    91.3  \\
         \citep{wang2022deepstruct} &           \shortstack{Supervised\\Zero-shot} &    Sent. &         No &    GLM &                          93 &                     87.8  \\
         \citep{liang2020bond} &                Distant Supervision &   Sent. &                         No & Roberta     &                                          91.21	& 86.2  \\
         \citep{cui2021template} &              Few-shot &   Sent. &                                    No & BART           &                       \shortstack{91.73\\92.55*} & -  \\
         \citep{luo2020hierarchical} &          Supervised &    Both &                                 No & BERT                                &                      93.37 &	90.3  \\
         \citep{yang2020simple} &               Few-shot &   Sent. &                                    No & BERT &                                 75.2* &                     -  \\
         \citep{lison2020named} &               Weak Supervision &    Doc. &                           No & BERT &                                                  71.6 &  -  \\
         \citep{wang2020pyramid} &              Supervised &   Sent. &                                  Yes & BERT    &                                 - &                     -  \\
         \citep{shen2021locate} &               Supervised &   Sent. &                                  Yes &    \shortstack{Glove\\BERT\\BIOwordvec} &                         - & -  \\
         \citep{li2022unified} &                Supervised &   Sent. &                                  Yes &    BERT &                                                93.07&	90.5 \\
         \citep{schweter2020flert} &            Supervised &   Doc. &                                  No & \shortstack{Roberta\\Glove}         &                           93.75 & -  \\
         \citep{wang2020automated} &            Supervised &  Sent. &                                   No &  Multiple                                    &                   94.6 & -  \\
             \citep{ye2021packed} &             Supervised &  Sent. &                                   Yes &  BERT variants                         &                   94.0 & 91.9  \\
         \citep{liu2022autoregressive} &        Supervised &    Sent. &                                 No &   T5                                  &                                - & - \\
         \citep{yang2024pner} & Supervised & Sent. & Yes & \shortstack{BERT+BiLSTM+CRF\\BERT} & - & - \\
         \citep{ma2023coarse} & Few-shot & Sent. & No & - & - & - \\
        \citep{mao2024simple} & Supervised & Sent. & Yes & \shortstack{BERT\\BiLSTM} & - & - \\
         
    \end{tabular}
    \caption{The NER models properties and performance. Results with * are for few-shot learning.}
    \label{tab:NER}
\end{table}

\section{Named Entity Recognition Models} \label{sec:NER}

This section covers the models that addressed the NER task. We show main NER models' properties in Table \ref{tab:NER}, which are: the model learning type, the used language model, the input level, such as sentence and document, and the ability to handle nested entities. Additionally, we show the models' F1 score on two common datasets, CoNLL2003 and OntoNotes5.0. Then we discuss the models' work below.

\subsection{Comprehensive NER Models}
Comprehensive NER models tackle both nested and flat entities. Machine Reading Comprehension (MRC) methods handled NLP problems as a question answering task. BERT-MRC \citep{li2019unified} targeted the different types of entities by extracting them from text through responding with answers to a query. For instance \textit{``Washington was born into slavery on the farm of James Burroughs"} which is an example given in their work, to extract the entity \textit{``Washington"}, thus, the query can be \textit{``which person is mentioned in the text?"}. Such approach supported extracting nested entities and utilized latent entity types in the query. On the other hand, the work in \citep{yu2020named} defined the NER task as the detection of the indexes of entity heads and entity tails in a sentence. Unlike state of the art works at the time of this publication, the model did not use lexical and syntactic (hand-crafted) features in the input, but utilized dependency parsing graph features in addition to the word representations generated by BERT \citep{devlin2018bert} and character representations. At the last stage, the Biaffine model \citep{dozat2016deep} was used to give scores in the output to determine the valid entities. The above models used several nested and flat NER datasets for evaluation. However, the latest showed better results in two of three nested NER datasets. The work in \citep{shen2021locate} considered the nested entities problem in an approach that is similar to object detection in the computer vision domain. For instance, in their given example, an object of a person may hold other sub-objects like a tennis racket or a hand watch. Thus, authors adopted the two-stage object detector algorithm and customized it for the NER task. In addition to using different PLMs for different datasets in the evaluation, features, such as part-of-speech tagging (POS) and character-level representation were employed. Pyramid \citep{wang2020pyramid} is a layered model that handled deep nested entities. The text input was represented on character and word levels and fed to an LSTM encoding layer, then multiple layers processed the input; each level had LSTM and CNN sub-components. The model showed significant performance on deep nested entities. For instance, the study showed an example of extracting eight nested entities from one term. Despite this, the model is still considered not easy for further enhancement or customization due to using several components. W$^2$NER model \citep{li2022unified} was designed to capture all types of entities: flat, nested and discontinuous. The model leveraged the relation between entity words to identify entity boundaries. Two types of relations were considered and used in a 2D matrix to find all the relations between all the word interactions within a sentence. However, such mechanism may incur additional computations when trying to identify $(n^2 - n)$ matches, where $n$ is the number of words in a sentence. The model used additional bidirectional LSTM layers to capture additional contextual information in the text encoding level. Additionally, multiple components were used to refine the results. Flat and nested NER datasets were used in the evaluation. The model in \citep{zheng2019boundary} combined two components in a multi-task learning model. The first used a sequence labelling layer to detect entity boundaries without the common error propagation problem. Whereas the second employed a region classification model to classify the entity boundaries. The evaluations used a biomedical datasets and German nested entities dataset. The model used character level representation for the input. Nevertheless, the results can be improved when leveraging other PLMs that have shown better scores in other tasks. The model in \citep{tan2021sequence} recognized nested entities by predicting a set under supervised learning. A sentence is encoded using a combination of Bert, Glove, part-of-speech tags and character level embeddings, then a non-autoregressive decoder makes predictions based on the number of predefined entities. To match the predictions with the gold entities a bipartite matching-based loss function was used. In the study conducted by Yang \citep{yang2024pner} the challenges of NER are discussed to be solved. It benefits from pipelining Some approaches including sequence labeling and text classification missions. Here to do the sequence labeling methods like BERT, and BiLSTM \citep{graves2013speech}, and CRF \citep{lafferty2001conditional}are implemented and for the text classification task BERT model is hired.
In \citep{mao2024simple}, a less studied problem in NERs related to the discontinuous NER is addressed. Hence, this model can be used not only for nested and flat NER but also for discontinuous NER. The mission here is to create the discontinues entities by putting unattached spans in it.    


\subsection{Flat NER Models}

This section surveys the models that did not address nested entities.

The model architecture in \citep{akbik2019pooled} handled the NER task by utilizing better text representation and employing contextualized character-level embeddings. Memory space was used to store the embeddings generated for each word. Employing memory storage implied the need to manage speed and capacity. However, such consideration was not discussed in the paper. Pooling operations were used to compute word embeddings based on the ones stored in the memory. TENER model \citep{yan2019tener} utilized character level encoding and adapted transformers' attention for efficient text context information capturing, thus, the model became aware of the distances between the words and the direction of context. FLERT \citep{schweter2020flert} is an extension of a previous model (FLAIR) \citep{akbik2019flair} which exploited document-level features for NER. Briefly the method employed two subsets of the text that surrounds a sentence in the input, and the output contained NER tags for the input sentence without the surrounding text. The implementation limited the surrounding tokens to 64 words before and after. The model in \citep{wang2021improving} addressed two types of the NER task that are: offline NER, where external resources can be used to enrich the input with related text. And the online NER, where cooperative learning minimized the distance between the input representation and the output distribution. Both NER types were handled in the proposed unified model.

Automated Concatenation of Embeddings (ACE) \citep{wang2020automated} proposed an approach for selecting the best combination of word representations for several tasks including NER. Authors employed reinforcement learning and proposed automated concatenation of embeddings. The work did not present an advanced model architecture for NER but utilized better word embeddings.

TriggerNER model \citep{lin2020triggerner} exploited words that surround an entity in a sentence to perform the NER task; authors named those surrounding words as entity triggers and by identifying patterns of triggers, they trained a sequence tagging model for the task. They benefited from crowd sourced annotated triggers in training a model that learned entity triggers, then the NER output model depended on the information from the first component.


The authors in \citep{liang2020bond} explained that traditional deep learning approaches require enormous training data, thus it is more theoretical than adjustable to real-world data. They proposed BOND, a distant supervision model, that utilized small amount of labeled samples to annotate large portion of the used datasets. They tackled two main issues in distant supervision learning, the incomplete annotation and the generated noise, by a two-stage training framework. They employed Roberta \citep{liu2019roberta} to generate labels, then used the labels in the second stage self-training. With additional training iterations the model achieved competitive results in the distant-supervision learning, the study also reported the fully-supervised learning performance. Nevertheless, the gap between the fully-supervised and the distant-supervised performances is still large; but using larger language models it could reduce that gap as stated in their conclusion.

\subsection{Document-level Models}

In \citep{luo2020hierarchical}, authors proposed a model for both sentence-level and document-level datasets; they employed label embeddings in the sentence level and used it to find a similarity score between each label and its input word. In the document level, a key-value memory was employed for all the embeddings used during training. The input consisted of word and character representations. In \citep{lison2020named}, a weak supervision model employed external knowledge to label data through several labelling functions derived from different models, such as sequence labelling and heuristic functions, then output items from the different functions are aggregated for the last sequence labelling step in the model.

The work in \citep{luo2020hierarchical} showed that recurrent neural network (RNN) layers, that are commonly used in the NER task, suffer from some limitations; specifically, long -short term memory (LSTM) layers do not handle sentence-level information as expected and they are not designed for document-level data by nature. Authors proposed a model that can handle sentence-level and document-level data. They used BERT for word-level representation and IntNet \citep{xin2018learning} for character-level representation in a hierarchical contextualized representation architecture . The model employed label embeddings to find the closest label for words in a sentence. In the document-level training, a key-value store memorized all the word representations to be used at once. Nevertheless, the model can witness better performance when using advanced memory handling algorithms for large scale datasets.

The work in \citep{lison2020named} handled only document-level data in weak supervision manner, thus, unlabelled data was used in training, which solves the problem of finding high quality labelled datasets for specific domains. However, all the datasets were based on news articles, thus, the model was not evaluated to generalize on various domains. In their model, multiple labelling function annotated the entities, then the output was aggregated; after that a function was trained to label the entities in the text sequence. Their word presentation was based on BERT and their model did not detect nested entities.


\subsection{Few-shot NER Models}



The work in \citep{cui2021template} used manually created templates of facts retrieved from datasets to train the model. For instance, \textit{``Bangkok is a location entity"} is a given template example that is retrieved from the fact ``ACL will be held in Bangkok". The model adapted easily on new domains with few samples by fine-tuning the original model. The results showed also the model's performance on supervised learning.

StructShot model \citep{yang2020simple} utilized contextual representations of the labels from the support set instead of the traditional approaches. To test the effectiveness of their approach, authors used general dataset in the source domain and tested the model on several datasets from other domains. They reported the performance on one-shot and five-short performances. The model experienced additional step of learning label representations in supervised training and did not detect nested entities.

ContaiNER \citep{das2021container} employed contrastive learning for the NER task by decreasing the distance between similar entities and increasing the distance between unsimilar ones, especially to differentiate between predefined entities and the entities that are categorized as not belonging to the predefined set, know as entities with the outside (O) tag.

The paper in \citep{hou2020few} proposed L-TapNet+CDT, a model that used conditional random fields (CRF) to exploit label dependencies from the source domain to the target domain in the few-shot scope. Additionally, the authors proposed L-TapNet to enlarge the gap between label embeddings. This approach reflected better classifications supported by the ability to detect the similarity between an input word and its label, such as ``rain" and ``weather".

MUCO model \citep{tong2021learning} exploited the words that belonged to the non-entity class (O-class) by clustering them in order to support entity words classification. In detail, a classifier was trained to learn to cluster entity pairs based on the non-entity class word that falls between any pair. Thus, the model explored common semantics between entities that belonged to the same cluster. The model was not evaluated on few-shot datasets, such as Few-NERD but split and customized some supervised datasets for the task.
 
MAML-ProtoNet model \citep{ma2022decomposed} consisted of two components to enhance entity span level tagging and to mitigate the effect of non-entity class (O-class) spans, especially because O-class spans do not provide much common information. The first component only detected spans without labeling them with any of the pre-defined classes, whereas the following component did the labeling. In such an approach, the non-entity class did not harm the first stage as labeling was not required.

The article in \citep{ma2023coarse} introduces C2FNER model with the goal of rapid adjustment to the new class of entity with minimal data. In this article, research is centered around training a model on a coarse-grained class and then employing the trained model to distinguish fine-grained class using Few-shot learning in NER. As an example, finding the sub-classes among the main class or moving from a general to a more detailed classification is the aim of this study.

\section{Relation Classification} \label{sec:relation}

RC models determine if a relation exists between two given entities and classify it into one of predefined relations. Our survey includes some few-shot models that were selected based on the criteria in Section \ref{sec:methodology}. Table \ref{tab:RC} summarizes the properties of the RC models. We show the machine learning approach. Furthermore, the input level addressed by the model. Also, the used language model. The last column is an indicator of the output format for RC and RE models. Table \ref{tab:RCresults} shows the reported F1 score when found for the considered model on two common datasets TACRED and FewRel. The last four columns show the FewRel F1 score on 5-way-1shot, 5way-5shot, 10way-1shot, and 10way-5shot performance respectively.

RECENT \citep{lyu2021relation} is an RC model-agnostic paradigm, that enhances the performance by restricting the candidate prediction relations based on the entity types. When applied to SpanBERT \citep{joshi2020spanbert}, the model achieved a new F1 score on the TACRED dataset.

TACNN \citep{geng2022novel} proposed a target attention mechanism which assigned increased weights to important entities in the sentence to enhance identifying a target relation. Although the study was published recently, several older models outperformed their reported F1 scores. TACNN did not utilize contemporary or contextualized language models, such as Bert and GPT-3 \citep{brown2020language}, but used Word2vec \citep{mikolov2013efficient}. Additionally, the word embeddings were extended by concatenating them to positional embeddings, then the attention technique is used followed by convolutional layers.

\subsection{Few-shot RC Models}

The work in \citep{xie2020heterogeneous} used a heterogeneous graph neural network (HGNN) for few-shot learning task of predicting the relation as a node classification problem. Entities and sentences represent different node types in the graph. Entity nodes fill the gap between the sentence node and the valid relation node. Adversarial learning was utilized to make the model robust to noisy data. The text was encoded using Glove PLM. However, the model followed a traditional approach to encode the nodes, instead of advanced graph embedding algorithms. The model was evaluated on FewRel 1.0 dataset.

Logic-guided Semantic Representation Learning (LSRL) \citep{li2020logic} is an approach that utilizes two types of features from knowledge graphs. First, entity and relation embeddings to identify connections between relations. Second, relation inferring rules using rule mining methods. The features are utilized along with the word representations to connect unseen relations to seen ones. The method is model-agnostic; it was evaluated on two zero-shot models, DeViSE \citep{frome2013devise} and ConSE \citep{norouzi2013zero}. The models were evaluated on a dataset that was constructed for this research from Wikipedia articles.

TD-Proto \citep{yang2020enhance} utilized relation and entity descriptions to enhance prototypical network-based model. Prototypical networks, finds a prototype for classes and sentences. These networks have been adopted by several RC models and reflected good performance as they supported matching queries with prototypes \citep{gao2019hybrid, ye2019multi}.

ProtoNet \citep{ren2020two} is a prototypical network-based model. Authors showed that few-shot learning models can handle real-world problems better when they leverage the massive training data that is available to use. At the same time, these few-shot learning models should handle novel relations. Thus, they combined prototypical techniques from supervised learning and few-shot learning. Furthermore, the used loss function targeted enlarging the distance between the relation representations in the embeddings space.


The work in \citep{peng2020learning} examined the contribution of the input components in the RE task, the text context and the entities. They performed experiments on datasets that are commonly used in the task to understand the effect of each component, and they showed that the currently used datasets do not support objective evaluation. Furthermore, they showed that there is still further information in the textual context to be absorbed by models to enhance the results. Based on that, authors proposed a training framework that tackled the mentioned findings by applying masks to portion of the entities.

\renewcommand{\arraystretch}{1.25}
\begin{table}[t]
    \centering
    \begin{tabular}{llcccccccc}
    \hline
         Model & \shortstack{Learning Type} & Sent./Doc. & \shortstack{Language Model} &  RE/RC \\
         \hline
         \citep{he2023virtual} &            Multi. &      Sent.     & GLM                   & RE  \\
         \citep{he2023virtual} &            Few-shot &      Sent.     & Custom                   & RE  \\
         \citep{chen2023dialogue} &        Supervised &         Doc.     & Glove                   & RE  \\
         \citep{sui2023joint} &             Supervised &    Sent.     & BERT                   & RE  \\
         \citep{ren2020two} &               Few-shot &      Sent.     & BERT                   & RE  \\
         \citep{xie2020heterogeneous} &        Few-shot &   Sent.     & Glove                   & RC  \\
         \citep{chen2022knowprompt} &        Supervised &   Sent.     & Roberta                   & RE  \\
         \citep{nan2020reasoning} &             Supervised & Doc.     & BERT                   & RE  \\
         \citep{zhong2020frustratingly} &        Supervised &  Sent.   & BERT                   & RE  \\
         \citep{guo2019attention} &         Supervised &      Both     & \shortstack{Graph\\encoding}                   & RE  \\

    \end{tabular}
    \caption{The RC models' properties.}
    \label{tab:RC}
\end{table}

Virtual prompt pre-training \citep{he2023virtual} is a few-shot learning model based on a novel prompt tuning approach. The work explained the prompt tuning as a new paradigm for training language models used in various tasks under the objective of predicting masked tokens. In this work, pre-training focused on detecting entities and relations. They used GLM \citep{du2021all} as the language model to encode text. The work was evaluated only on the two versions of FewRel.

Unlike several works that focused on sentences and other ones for documents, DHGAT \citep{chen2023dialogue} is a relation extraction model for dialog-type input, dialog datasets add extra difficulty due to the causality and less structured text used in it. The model encoded text using Glove \citep{pennington2014glove} in addition to part-of-speech tagging and entity type features in the input. The model used heterogeneous graph attention network to train the model, the graph contained multiple node types, such as utterance nodes, type nodes, word nodes, speaker nodes, and argument nodes.

ProtoNet \citep{ren2020two} is an incremental few-shot leaning model that benefited from existence of large-scale datasets to train the model on the existing relations then applied few-shot learning for the novel relations. Authors used prototype attention alignment to reduce the gap between the learned relations embeddings and the novel relations. The model was tested on FewRel 1.0 dataset.

Knowprompt \citep{chen2022knowprompt} is a supervised model that targeted enhancing the word representation by using prompt-tuning. They tackled some challenges in prompt-tuning through enriching the process with extra knowledge. For instance, the model provided entity types during fine-tuning the language model. The model encoded the input using Roberta PLM and promises better results if employed other PLM that appeared after the release of the model. The approach was test on several known RE datasets. However, the experienced complexity due to the usage of several sub-components which may make hard to accept customization or expand to other domains.

Attention Guided Graph Convolutional Networks (AGGCN) model \citep{guo2019attention} is based on dependency parsing graphs. The model enhanced the utilization of information in dependency parsing through graph \textit{soft prunning}. The model operated on cross-sentence and single sentence levels. Nevertheless, word embeddings have proved representing more powerful text information, thus the graph embedding could be enhanced by including word embeddings in the input encoder level.

\renewcommand{\arraystretch}{1.25}
\begin{table}[t]
    \centering
    \begin{tabular}{lccccccc}
    \hline
         Model & TACRED & FewRel & F-5W1S & F-5W5S & F-10W1S & F-10W5S \\
         \hline
         \citep{wang2022deepstruct}     &   76.8	& 1.0 &	98.4 &	100 &	97.8 &	99.8 \\
         \citep{he2023virtual}     &   -	& 2.0 &	95.32 & 97.84 &	90.08 &	95.96\\
         \citep{ren2020two}     &   - & 1.0 &	82.1 &	84.64 &	- &	- \\
         \citep{xie2020heterogeneous}     &   -	& 1.0 &	73.83	& 87.12 &	62.15 &	74.23 \\
         \citep{chen2022knowprompt}     &   72.4	&	- &- &	- &	- &	- \\
         \citep{guo2019attention}     &   69.0	&	- &- &	- &	- &	- \\

    \end{tabular}
    \caption{The RC models' performance. The FewRel column shows the dataset version.}
    \label{tab:RCresults}
\end{table}

Latent Structure Refinement (LSR) \citep{nan2020reasoning} generated task-specific dependency graph structures for document level relations. The mode performed on the supervised learning paradigm. The model used iterative refinement during training to build global interactions knowledge. Text encoder, such as BERT, was used to generate token representation, then entity representation is used as nodes in the constructed graph in addition to nodes that reflected tokens dependency. The model was evaluated using DocRED \citep{yao2019docred} dataset only, probably due the lack of document-level data. However, it was compared to various baseline models with different architecture and showed superiority.

\section{Conclusion} \label{sec:conclusion}

We present a survey of recent deep learning models that address named entity recognition and relation classification, with focus on few-shot learning performance. In named entity recognition models, we find that entity boundary issue should be handled in the coming works since considering partial match as a correct prediction in multi-word entities is not a trusted evaluation. Furthermore, we find that models can benefit from the advances in language models' prompt-tuning to build strong architectures to achieve new state-of-the-art scores, since current models either focus on proposing a complicated model design, or focus on enhancing the word representation.

In the relation classification task, researchers should direct their efforts towards cross-sentence or document level achievements under the few-shot learning discipline, since this reflects more realist scenarios. Furthermore, there is lack in datasets for evaluating such type of work. Additionally, efforts should consider combining linguistic features with dependency parsing information to support the reliance on language models and score new results.




\bibliographystyle{ACM-Reference-Format}
\bibliography{ref}

\end{document}